\documentclass[11pt]{article}
\usepackage{sibarticle}
\usepackage{lmodern}
\usepackage{url}            
\usepackage{nicefrac}       
\usepackage{microtype}      

\usepackage{amssymb,graphicx}
\usepackage{float,amsmath}
\usepackage{amsfonts,epsfig}
\usepackage[inoutnumbered,linesnumbered,algoruled,slide,vlined]{algorithm2e}
\usepackage{subcaption}
\usepackage{color}
\usepackage[framemethod=tikz]{mdframed}
\usepackage{xcolor}
\usepackage{xparse}
\usepackage{xstring}

\title{Bolstering Stochastic Gradient Descent with Model Building} 
\ShortTitle{Stochastic Model Building (SMB)}
\ShortAuthors{Birbil, Martin, Onay \& Öztoprak}

\NumberOfAuthors{4}
\FirstAuthor{Ş. İlker Birbil}
\FirstAuthorAddress{University of
Amsterdam, 11018 TV Amsterdam, The Netherlands}

\SecondAuthor{Özgür Martin}
\SecondAuthorAddress{Mimar Sinan Fine Arts University, 
34380 Istanbul, Turkey}

\ThirdAuthor{Gönenç Onay}
\ThirdAuthorAddress{Galatasaray University, 34349 Istanbul, Turkey \\[1.2mm]
Coach-Ai GmbH - AI \& Analytics, 64295 Darmstadt, Germany 
}

\FourthAuthor{Figen Öztoprak}
\FourthAuthorAddress{Gebze Technical University,
41500 Kocaeli, Turkey}


\allowdisplaybreaks

\begin{document}

\maketitle
\sloppy
\begin{abstract}
	Stochastic gradient descent method and its variants constitute the core optimization algorithms
	that achieve good convergence rates for solving machine learning problems.
	These rates are obtained especially when these algorithms are fine-tuned for the application at hand.
	Although this tuning process can require large computational costs, recent work has shown that these costs can be reduced by line search methods that iteratively adjust the step length. We propose an alternative approach to stochastic line search by using a new algorithm based on forward step model building. This model building step incorporates second-order information that allows adjusting not only the step length but also the search direction. Noting that deep learning model parameters come in groups (layers of tensors), our method builds its model and calculates a new step for each parameter group. This novel diagonalization approach makes the selected step lengths adaptive. We provide convergence rate analysis, and experimentally show that the proposed algorithm achieves faster convergence and better generalization in well-known test problems. More precisely, SMB requires less tuning, and shows comparable performance to other adaptive methods.
\end{abstract}


Stochastic gradient descent (SGD) is a stochastic-approximation type optimization algorithm with several variants and a well-studied theory \citep{tadic:1997, chen:2023}.  It is a popular choice for machine learning applications; in practice, it can achieve fast convergence when its stepsize and its scheduling are tuned well for the specific application at hand. However, this tuning procedure can take up to thousands of CPU/GPU days resulting in big energy costs \citep{Asi2019}.  A number of researchers have studied adaptive strategies for improving the direction and the step length choices of the stochastic gradient descent algorithm. Adaptive sample size selection ideas \citep{Byrd:2012,Balles:2016,Bolla:2018} improve the direction by reducing its variance around the negative gradient of the empirical loss function, while stochastic quasi-Newton algorithms \citep{Byrd:2016,Wang:2017} provide adaptive preconditioning. Recently, several stochastic line search approaches have been proposed. Not surprisingly, some of these work cover sample size selection as a component of the proposed line search algorithms \citep{Balles:2016, Paq:2020}.

The Stochastic Model Building (SMB) algorithm proposed in this paper is not designed as a stochastic quasi-Newton algorithm in the sense explained by \cite{Bottou:2018}. However, it still produces a scaling matrix in the process of generating trial points, and its overall step at each outer iteration can be written in the form of matrix-vector multiplication. Unlike the algorithms proposed by \cite{Mokhtari:2014} and \cite{Schrau:2007}, we have no accumulation of curvature pairs throughout several iterations. Since there is no memory carried from earlier iterations, the scaling matrices in individual past iterations are based only on the data samples employed in those iterations. In other words, the scaling matrix and the incumbent random gradient vector are dependent. That being said, we also provide a version (SMBi), where the matrix and gradient vector in question become independent (see Algorithm \ref{alg:SMBi}).

\cite{Vas:2019} apply a deterministic globalization procedure on mini-batch loss functions. That is, the same sample is used in all function and gradient evaluations needed to apply the line search procedure at a given iteration. However, unlike our case, they employ a standard line search procedure that does not alter the search direction. They establish convergence guarantees for the empirical loss function under the \emph{interpolation} assumption, which requires each component loss function to have zero gradient at a minimizer of the empirical loss. \cite{Muts:2020} assume that the optimal learning rate (\textit{i.e.}, step length) along the negative batch gradient is a good estimator for the optimal learning rate with respect to the empirical loss along the same direction. They test validity of this assumption empirically on deep neural networks (DNNs). Rather than making such strong assumptions, we stick to the general theory for stochastic quasi-Newton methods.

Other work follow a different approach to translate deterministic line search procedures into a stochastic setting, and they do not employ fixed samples. In \cite{Hennig:2017}, a probabilistic model along the search direction is constructed via techniques from Bayesian optimization. Learning rates are chosen to maximize the expected improvement with respect to this model and the probability of satisfying Wolfe conditions.  \cite{Paq:2020} suggest an algorithm closer to the deterministic counterpart, where the convergence is based on the requirement that the stochastic function and gradient evaluations approximate their true values with a high enough probability.

Finally, we should mention that the finite-sum minimization problem is a special case of the general expected value minimization problem, for which certain modification ideas for SGD regarding the selection of the search direction and the step length can be applicable.  One such idea is \emph{gradient aggregation}, which adds to the search direction of SGD a variance reducing component obtained via stochastic gradient evaluations at previous iterates \citep{roux:2012, defazio:2014}.  In \cite{malinovsky:2022}, an aggregated-gradient-type step is produced in a distributed setting where the overall step is produced by employing step lengths at two levels.  Another idea is to use an extended step length control strategy depending on the objective value and the norm of the computed direction that might occasionally set the step length to zero \citep{liuzzi:2022}.  However, it is not clear how these ideas can be extended to the more general case of expected value minimization. 


With our current work, we make the following contributions. We use a model building strategy for adjusting the step length and the direction of a stochastic gradient vector. This approach also permits us to work on subsets of parameters. This feature makes our model steps not only adaptive, but also suitable to incorporate into the existing implementations of DNNs. Our method changes the direction of the step as well as its length. This property separates our approach from the backtracking line search algorithms. It also incorporates the most recent curvature information from the current point. This is in contrast with the stochastic quasi-Newton methods which use the information from the previous steps. Capitalizing our discussion on the independence of the sample batches, we also give a convergence analysis for SMB. Finally, we illustrate the computational performance of our method with a set of numerical experiments and compare the results against those obtained with other well-known methods.

\section{Stochastic Model Building.} \label{sec:SMB}

We introduce a new stochastic unconstrained optimization algorithm in order to approximately solve problems of the form
\begin{equation}
	\label{eq:origprob}
	\min_{x\in\Re^n} \ \ f(x) = \mathbb{E}[F(x, \xi)],
\end{equation}
where $F: \mathbb{R}^n \times \mathbb{R}^d \to \mathbb{R}$ is continuously differentiable and possibly nonconvex, $\xi \in \mathbb{R}^d$ denotes a random variable, and $\mathbb{E}[.]$ stands for the expectation taken with respect to $\xi$. We assume the existence of a stochastic first-order oracle which outputs a stochastic gradient $g(x, \xi)$ of $f$ for a given $x$. A common approach to tackle (\ref{eq:origprob}) is to solve the empirical risk problem
\begin{equation}
	\label{eq:erm}
	\min_{x\in\Re^n} \ \ f(x) =\frac{1}{N}\sum_{i=1}^{N} f_i(x),
\end{equation}
where  $f_i:\mathbb{R}^n \to \mathbb{R}$ is the loss function corresponding to the $i$th data sample, and $N$ denotes the data sample size which can be very large in modern applications.

As an alternative approach to line search for SGD, we propose a stochastic model building strategy inspired by the work of \cite{Oztoprak:2017}. Unlike core SGD methods, our approach aims at including a curvature information that adjusts not only the step length but also the search direction. \cite{Oztoprak:2017} consider only the deterministic setting and they apply the model building strategy repetitively until a sufficient descent is achieved. In our stochastic setting, however, we have observed experimentally that using multiple model steps does not benefit much to the performance, and its cost to the runtime can be extremely high in large-scale (\textit{e.g.}, deep learning) problems. Therefore, if the sufficient descent is not achieved by the stochastic gradient step, then we construct only one model to adjust the length and the direction of the step.

Conventional stochastic quasi-Newton methods adjust the gradient direction by a scaling matrix that is constructed by the information from the previous steps. Our model building approach, however, uses the most recent curvature information around the latest iteration. In popular deep learning model implementations, model parameters come in groups and updates are applied to each parameter group separately. Therefore, we also propose to build a model for each parameter group separately making the step lengths adaptive.

The proposed iterative algorithm SMB works as follows: At step $k$, given the iterate $x_k$, we calculate the stochastic function value $f_k = f(x_k,  \xi_{k})$ and the mini-batch stochastic gradient $g_k = \frac{1}{m_k}\sum_{i=1}^{m_k}g(x_k, \xi_{k,i})$ at $x_k$, where $m_k$ is the batch size, and $\xi_k = (\xi_{k,1},\ldots,\xi_{k,m_k})$ is the realization of the random vector $\xi$. Then, we apply the SGD update to calculate the trial step $s_k^t = - \alpha_k g_k$, where $\{\alpha_k\}_k$ is a sequence of learning rates. With this trial step, we also calculate the function and gradient values $f^t_k = f(x^t_k,  \xi_{k})$ and $g^t_k = g(x^t_k, \xi_{k})$ at $x^t_k = x_k + s^t_k$. Then, we check the stochastic Armijo condition
\begin{equation}
	\label{ineq:armijo}
	f^t_k  \leq f_k - c \ \alpha_k \|g_k\|^2,
\end{equation}
where $c > 0$ is a hyper-parameter. If the condition is satisfied and we achieve \textit{sufficient decrease}, then we set $x_{k+1} = x^t_k$ as the next step. If the Armijo condition is not satisfied, following \cite{Oztoprak:2017}, we build a quadratic model using the linear models at the points $x_{k,p}$ and $x^t_{k,p}$ for each parameter group $p$ and find the step $s_{k,p}$ to reach its minimum point.  Here, $x_{k,p}$ and $x^t_{k,p}$ denote respectively the coordinates of $x_{k}$ and $x^t_{k}$ that correspond to the parameter group $p$. We calculate the next iterate $x_{k+1} = x_k + s_k$,  where $s_k = (s_{k,p_1}, \ldots, s_{k,p_r})$ and $r$ is the number of parameter groups, and proceed to the next step with $x_{k+1}$ . This model step, if needed, requires extra mini-batch function and gradient evaluations (forward and backward pass in deep neural networks).

\begin{figure}[H]
	\centering
	\includegraphics[width=0.5\textwidth]{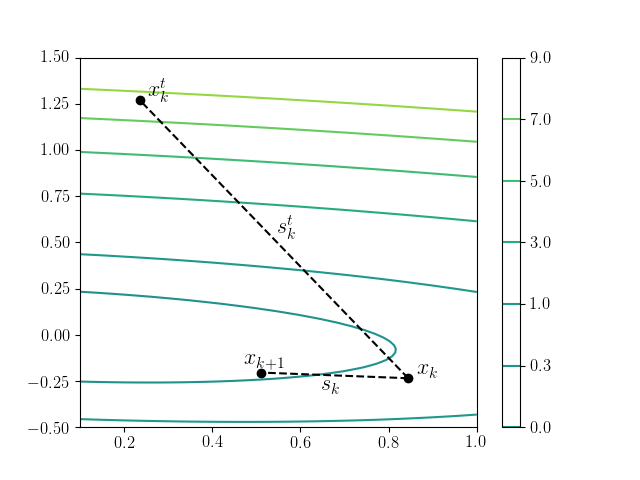}
	\caption{An iteration of SMB on a simple quadratic function. We assume for simplicity that there is only one parameter group, and hence, we drop the subscript $p$ . The algorithm first computes the trial point $x_k^t$ by taking the (stochastic) gradient step $s_k^t$. If this point is not acceptable, then it builds a model using the information at $x_k$ and $x_k^t$, and computes the next iterate $x_{k+1}=x_k+s_k$. Note that $s_k$ not only have a smaller length compared to the trial step $s_k^t$, but it also lies along a direction decreasing the function value.}
	\label{fig:describe}
\end{figure}

For each parameter group $p \in \{p_1, \ldots, p_r\}$, the quadratic model is built by combining the linear models at $x_{k,p}$ and $x^t_{k,p}$, given by
$$l_{k,p}^0(s) := f_{k} + g_{k,p}^\top s \ \ \ \mbox{ and } \ \ \ l_{k,p}^t(s-s^t_{k,p}) := f^t_{k} + (g^t_{k,p})^\top (s-s^t_{k,p}),$$
respectively. Then, the quadratic model becomes
$$m^t_{k,p}(s) = \alpha_{k,p}\ell^0_{k,p} + (1 - \alpha_{k,p})\ell^t_{k,p},$$
where
\begin{equation*}
	\alpha_{k,p} = -\frac{(s - s^t_{k,p})^\top s^t_{k,p}}{\|s^t_{k,p}\|^2}.
\end{equation*}
The constraint
$$ \|s\|^2 + \|s-s^t_{k,p}\|^2 \leq \|s^t_{k,p}\|^2,$$
is also imposed so that the minimum is attained in the region bounded by $x_{k,p}$ and $x^t_{k,p}$. This constraint acts like a trust region. Figure \ref{fig:describe} shows the steps of this construction.

In this work, we solve a relaxation of this constrained model as explained in \cite[Section 2.2]{Oztoprak:2017} where one can find the full approach
for finding the approximate solution of the constrained problem. The minimum value of the relaxed model is attained at the point $x_{k,p} + s_{k,p}$ with
\begin{equation}
	\label{step_dkt}
	s_{k,p} = c_{g,p} (\delta) g_{k,p} + c_{y,p} (\delta) y_{k,p} + c_{s,p} (\delta) s^t_{k,p},
\end{equation}
where $y_{k,p} := g^t_{k,p} - g_{k,p}$.
Here, the coefficients are given as
\[
	c_{g,p} (\delta) = -\frac{\|s_{k,p}^t\|^2}{\delta}, \quad c_{y,p} (\delta) =
	-\frac{\|s_{k,p}^t\|^2}{\delta\theta}[-(y_{k,p}^\top s_{k,p}^t +
	\delta)(s_{k,p}^t)^\top g_{k,p} + \|s_{k,p}^t\|^2 y_{k,p}^\top g_{k,p}],
\]
\[
	c_{s,p}(\delta) = -\frac{\|s_{k,p}^t\|^2}{\delta\theta}[-(y_{k,p}^\top s_{k,p}^t + \delta)y_{k,p}^\top g_{k,p} + \|y_{k,p}\|^2(s_{k,p}^t)^\top g_{k,p}],
\]
with

\begin{equation}
	{\small
	\label{theta_delta}
	\theta = \left(y_{k,p}^\top s_{k,p}^t + 2\delta\right)^2-\|s_{k,p}^t\|^2\|y_{k,p}\|^2 \ \mbox{\normalsize and } \
	\delta =  \|s_{k,p}^t\|\left(\|y_{k,p}\|+\frac{1}{\eta}\|g_{k,p}\|\right) - y_{k,p}^\top s_{k,p}^t,}
\end{equation}
where $0 < \eta < 1$ is a constant. Then, the adaptive model step becomes $s_k = (s_{k,p_1}, \ldots, s_{k,p_r})$. We note that our construction in terms of different parameter groups lends itself to constructing a different model for each parameter subspace.

We summarize the steps of SMB in Algorithm \ref{alg:SMBcompact}. Line \ref{ln:xkt} shows the trial point, which is obtained with the standard stochastic gradient step. If this step satisfies the stochastic Armijo condition, then we proceed with the next iteration (line \ref{ln:noarm}). Otherwise, we continue with bulding the models for each parameter group (lines \ref{ln:forst}- \ref{ln:forend}), and move to the next iteration with the model building step in line \ref{ln:mbstep}.

\begin{algorithm}
	\SetArgSty{textit}
	\KwIn{ $x_1 \in \mathbb{R}^n$, step lengths $\{\alpha_k\}_{k=1}^T$, mini-batch sizes $\{m_k\}_{k=1}^T$, and $c > 0$}
	\For{$k=1,\ldots,T$}{
	$f_k = f(x_k, \xi_{k})$, $g_k =  \frac{1}{m_k}\sum_{i=1}^{m_k}g(x_k, \xi_{k,i})$\;
	$s^t_k = -\alpha_k g_k$\;
	$x_k^t= x_k + s^t_k $\; \label{ln:xkt}
	$f_k^t = f(x_k^t, \xi_{k})$\;
	\eIf{$f^t_k  \leq f_k - c \ \alpha_k \|g_k\|^2$}{
	$x_{k+1} = x^t_k$ \; \label{ln:noarm}
	}{$g_k^t = \frac{1}{m_k}\sum_{i=1}^{m_k}g(x^t_k, \xi_{k,i})$\;
	\For{\em $p \in \{p_1, \ldots, p_r\}$}{ \label{ln:forst}
	$y_{k,p} = g_{k,p} ^t - g_{k,p} $\;
	$s_{k,p}  = c_{g,p}(\delta) g_{k,p}  + c_{y,p}(\delta) y_{k,p}  + c_{s,p}(\delta) s^t_{k,p} $\; \label{ln:forend}
	}
	$x_{k+1} = x_k + s_k$ with $s_k = (s_{k,p_1}, \ldots, s_{k,p_r})$\; \label{ln:mbstep}
	}

	}
	\caption{SMB: Stochastic Model Building}
	\label{alg:SMBcompact}
\end{algorithm}

\paragraph{An example run.} It is not hard to see that SGD corresponds to steps 3-5 of Algorithm~\ref{alg:SMBcompact}, and the SMB step can possibly reduce to an SGD step.  Moreover, the SMB steps produced by Algorithm~\ref{alg:SMBcompact} always lie in the span of the two stochastic gradients, $g_k$ and $g_k^t$. In particular, when a model step is computed in line 13, we have
\[
	s_{k}=w_1g_k+w_2g_k^t \text{ with } w_1 = c_g(\delta)-c_y(\delta)-c_s(\delta)\alpha \text{ and } w_2 = c_y(\delta),
\]
where $\alpha$ is a constant step length. Therefore, it is interesting to observe how the values of $w_1$ and $w_2$ evolve during the course of an SMB run, and how the resulting performance compares to taking SGD steps with various step lengths. For this purpose, we investigate the steps of SMB for one epoch on the MNIST dataset with a batch size of 128 (see Section~\ref{sec:exp} for details of the experimental setting).

\begin{figure}[H]
	\centering
	\includegraphics[width=\textwidth]{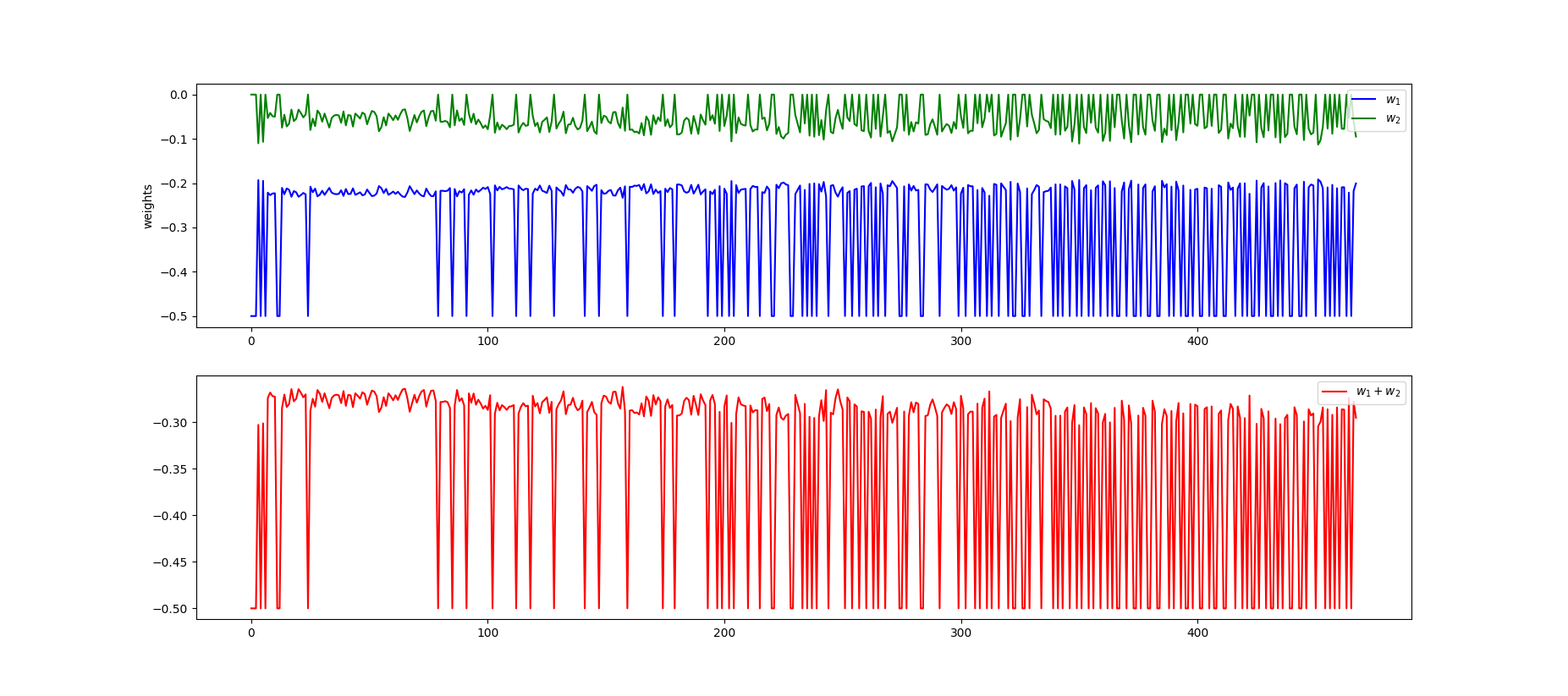}
	\caption{The coefficients of $g_k$ and $g_k^t$ during a single-epoch run of SMB on the MNIST data with $\alpha=0.5$. Model steps are taken quite often, but not at all iterations. The sum of the two coefficients vary in [-0.5,-0.25].}
	\label{fig:example}
\end{figure}

We provide in Figure~\ref{fig:example} the values of $w_1$ and $w_2$ for SMB with $\alpha=0.5$ over the 468 steps taken in an epoch. Note that the computations of $g_k^t$ in line 6 of Algorithm~\ref{alg:SMBcompact} may spend a significant portion of the evaluation budget, if model steps are taken very often. Figure~\ref{fig:example} shows that SMB algorithm indeed takes too many model steps in this run as indicated by the frequency of positive $w_2$ values. To account for the extra gradient evaluations in computing the model steps, we run SGD with a constant learning rate of $\alpha$ on the same problem for two epochs rather than one. (The elapsed time of sequential runs on a PC with 8GB RAM vary in 8-9 seconds for SGD, and in 11-15 seconds for SMB). Table~\ref{tab:example} presents a summary of the resulting training error and testing accuracy values.  We observe that the performance of SMB is significantly more stable for different $\alpha$ values, thanks to the adaptive step length (and the modified search direction) provided by SMB. SGD \emph{can} achieve performance values comparable to or even better than SMB, but only for the \emph{right} values of $\alpha$.  
In Figure~\ref{fig:example}, it is interesting to see that the values of $w_2$ are relatively small. We also realize that if we run SGD with a learning rate close to the average $(w_1+w_2)$ value, it has an inferior performance. For the SMB run with $\alpha=0.5$, for instance, the average $(w_1+w_2)$ value is close to $-0.3$. This can be contrasted with the resulting performance of SGD with $\alpha=0.3$ in Table~\ref{tab:example}. These observations suggest that $g_k^t$ contributes to altering the search direction as intended, rather than acting as an \emph{additional stochastic gradient step}.

\begin{table}[h]
	\centering
	\begin{tabular}{c | c c | c c | c c | c c | c c}
		              & \multicolumn{2}{c|}{$\alpha=1.0$} & \multicolumn{2}{c|}{$\alpha=0.5$} & \multicolumn{2}{c|}{$\alpha=0.3$} & \multicolumn{2}{c|}{$\alpha=0.1$} & \multicolumn{2}{c}{$\alpha=0.05$}                                              \\
		\hline
		              & SGD                               & SMB                               & SGD                               & SMB                               & SGD                               & SMB    & SGD    & SMB    & SGD    & SMB    \\
		\hline
		Training loss & 2.3033                            & 0.3402                            & 2.2947                            & 0.1770                            & 0.7435                            & 0.1889 & 0.1594 & 0.3379 & 0.2410 & 0.3131 \\
		Test accuracy & 0.1135                            & 0.8949                            & 0.1137                            & 0.9460                            & 0.7685                            & 0.9422 & 0.9513 & 0.8993 & 0.9298 & 0.9162 \\
	\end{tabular}
	\caption{Performance on the MNIST data; SMB is run for one epoch, and SGD is run for two epochs.}
	\label{tab:example}
\end{table}

\clearpage

\section{Convergence Analysis.} \label{sec:conv}

The steps of SMB  can be considered as a special quasi-Newton update:
\begin{equation}
	x_{k+1} = x_k -\alpha_k H_k g_k,
\end{equation}
where $H_k$ is a symmetric positive definite matrix as an approximation to the inverse Hessian matrix. In Appendix \ref{app:proof}, we explain this connection and give an explicit formula for the matrix $H_k$. We also prove that there exists $\underline{\kappa}, \overline{\kappa} > 0$ such that for all $k$, the matrix $H_k$  satisfies
\begin{equation}\label{eq:positivedefinite}
	\underline{\kappa} I \preceq H_k \preceq  \overline{\kappa} I,
\end{equation}
where for two matrices $A$ and $B$, $A \preceq B$ means $B - A$ is positive semidefinite. It is important to note that $H_k$ is built with the information collected around $x_k$, particularly, $g_k$. Therefore, unlike stochastic quasi-Newton methods, $H_k$ is correlated with $g_k$, and hence, $\mathbb{E}_{\xi_k}[H_k g_k]$ is very difficult to analyze. Unfortunately, this difficulty prevents us from using the general framework given by \cite{Wang:2017}.

To overcome  this difficulty and carry on with the convergence analysis, we modify Algorithm \ref{alg:SMBcompact} such that $H_k$ is calculated with a new independent mini batch, and therefore, it is independent of $g_k$. By doing so, we still build a model using the information around $x_k$. Assuming that $g_k$ is an unbiased estimator of $\nabla f$, we conclude that $\mathbb{E}_{\xi_k}[H_kg_k] = H_k \nabla f$. In the rest of this section, we provide a convergence analysis for this modified algorithm which we will call as SMBi (`i' stands for independent batch). The steps of SMBi are given in Algorithm \ref{alg:SMBi}. As Step 11 shows, we obtain the model building step with a new random batch.

\begin{algorithm}[H]
	\label{algorithm:SMBi}
	\SetArgSty{textit}
	\KwIn{ $x_1 \in \mathbb{R}^n$, step lengths $\{\alpha_k\}_{k=1}^T$, mini-batch sizes $\{m_k\}_{k=1}^T$, and $c > 0$}
	\For{$k=1,\ldots,T$}{
	$f_k = f(x_k, \xi_{k})$, $g_k =  \frac{1}{m_k}\sum_{i=1}^{m_k}g(x_k, \xi_{k,i})$\;
	$s^t_k = -\alpha_k g_k$\;
	$x_k^t= x_k + s^t_k $\;
	$f_k^t = f(x_k^t, \xi_{k})$\;
	\eIf{$f^t_k  \leq f_k - c \ \alpha_k \|g_k\|^2$}{
	$x_{k+1} = x^t_k$ \;
	}{
	\For{\em $p = 1, \dots, n$}{
	Choose a new independent random batch $\xi'_k$\;
	$g'_k =  \frac{1}{m_k}\sum_{i=1}^{m_k}g(x_k, \xi'_{k,i})$\;
	$(s^t_k)' = -\alpha_k g'_k$, $(x_k^t)'= x_k + (s^t_k)'$\;
	$(g_k^t)' = \frac{1}{m_k}\sum_{i=1}^{m_k}g((x^t_k)', \xi'_{k,i})$, $y'_{k,p} = (g_{k,p}^t)' - g'_{k,p} $\;
	$s_{k,p}  = -\alpha_k H'_{k,p} g_k$, where $H'_{k,p}$ is calculated using $g'_k$ and $y'_k$ as defined in Appendix\;
	}
	$x_{k+1} = x_k + s_k$ with $s_k = (s_{k,1}, \ldots, s_{k,n})$\;
	}

	}
	\caption{SMBi:  $H_k$ with an independent batch}
	\label{alg:SMBi}
\end{algorithm}

\noindent Before providing the analysis, let us make the following assumptions:

\noindent\textbf{Assumption 1:} Assume that $f: \mathbb{R}^n \to \mathbb{R}$ is continuously differentiable, lower bounded by $f^{low}$, and there exists $L  > 0$ such that for any $x,y \in \mathbb{R}^n$, $\|\nabla f(x) - \nabla f(y)\| \leq L \|x-y\|$.

\noindent\textbf{Assumption 2:} Assume that $\xi_k$, $k \geq 1$, are independent samples and for any iteration $k$, $\xi_k$ is independent of $\{x_j\}_{j=1}^k$, $\mathbb{E}_{\xi_k}[g(x_k, \xi_k)] = \nabla f(x_k)$ and $\mathbb{E}_{\xi_k}[\|g(x_k, \xi_k) - \nabla f(x_k)\|^2] \leq M^2$, for some $M > 0$.

Although Assumption 1 is standard among the stochastic unconstrained optimization literature, one can find different variants of the Assumption 2 (see \cite{khaled:2020} for an overview). In this paper, we follow the framework of \cite{Wang:2017} which is a special case of \cite{Bottou:2018}.

In order to be in line with practical implementations and with our experiments, we first provide an analysis covering the constant step length case for (possibly) non-convex objective functions. Below, we denote by $\xi_{[T]} = (\xi_1, \ldots, \xi_T)$ the random samplings in the first $T$ iterations. Let $\alpha_{max}$ be the maximum step length that is allowed in the implementation of SMBi with
\begin{equation}\label{eq:alphamax}
	\alpha_{max} \geq \frac{-1 + \sqrt{1+16\eta^2}}{4L\eta},
\end{equation}
where $0 < \eta < 1$. This hyper-parameter of maximum step length is needed in the theoretical results. Observe that since $\eta^{-1} > 1$, assuming $L \geq 1$ implies that it suffices to choose $\alpha_{max} \geq 1$ to satisfying (\ref{eq:alphamax}). This implies further that $2/(L\eta^{-1} + 2L^2\alpha_{max}) \leq \alpha_{max}$. The proof of the next convergence result is given in Appendix \ref{app:proof}.

\begin{theorem} \label{thm:wang}
	Suppose that Assumption 1 and Assumption 2 hold  and $\{x_k\}$ is generated by SMBi as given in Algorithm \ref{alg:SMBi}. Suppose also that $\{\alpha_k\}$ in Algorithm \ref{alg:SMBi} satisfies that $0 < \alpha_k < 2/(L\eta^{-1} + 2L^2\alpha_{max}) \leq \alpha_{max}$ for all $k$. For given $T$, let $R$ be a random variable with the probability mass function
	$$\mathbb{P}_R(k) := \mathbb{P}\{R=k\} = \frac{\alpha_k / (\eta^{-1} + 2L\alpha_{max})  - \alpha^2_kL / 2}{\sum_{k=1}^T (\alpha_k / (\eta^{-1} + 2L\alpha_{max})  - \alpha^2_kL / 2)}$$
	for $k = 1, \ldots, T$. Then, we have
	$$\mathbb{E}[\|\nabla f(x_R)\|^2] \leq \frac{D_f + (M^2L/ 2) \sum_{k=1}^T (\alpha_k^2/m_k)}{\sum_{k=1}^T (\alpha_k / (\eta^{-1} + 2L\alpha_{max})  - \alpha^2_kL / 2)},$$
	where $D_f := f(x_1) - f^{low}$ and the expectation is taken with respect to $R$ and $\xi_{[T]}$. Moreover, if we choose $\alpha_k = 1/(L\eta^{-1} + 2L^2\alpha_{max})$ and $m_k = m$ for all $k = 1, \ldots, T$, then this reduces to
	$$\mathbb{E}[\|\nabla f(x_R)\|^2] \leq \frac{2L(\eta^{-1}  + 2L\alpha_{max})^2 D
			_f}{T} + \frac{M^2}{m}.$$
\end{theorem}

Using this theorem, it is possible to deduce that stochastic first-order oracle complexity of SMB with random output and constant step length is $\mathcal{O}(\epsilon^{-2})$ \cite[Corollary 2.12]{Wang:2017}. In \cite{Wang:2017} (Theorem 2.5), it is shown that under our assumptions above and the extra assumptions of $0 < \alpha_k \leq  \frac{1}{L (\eta^{-1} + 2L\alpha_{max})} \leq \alpha_{max}$, $\sum_{k=1}^{\infty} \alpha_k = \infty$ and $\sum_{k=1}^{\infty} \alpha_k^2 < \infty$, if the point sequence $\{x_k\}$ is generated by SMBi method (when $H_k$ is calculated by an independent batch in each step) with batch size $m_k = m$ for all $k$, then there exists a positive constant  $M_f$ such that $\mathbb{E}[f(x_k)] \leq M_f$. Using this observation, the proofs of Theorem \ref{thm:wang},  and Theorem 2.8 in \citep{Wang:2017}, we can also give the following complexity result when the step length sequence is diminishing.

\begin{theorem}
	Suppose that Assumption 1 and Assumption 2 hold. Let the batch size \(m_k = m\) for all \(k\) and assume that $\alpha_k = \frac{1}{L (\eta^{-1} + 2L\alpha_{max})} k^{-\phi}$ with $\phi \in (0.5, 1)$ for all $k$. Then $\{x_k\}$ generated by SMBi satisfies
	$$\frac{1}{T} \sum_{k=1}^T \mathbb{E}[\|\nabla f(x_k)\|^2] \leq 2L (\eta^{-1} + 2L\alpha_{max})(M_f - f^{low}) T^{\phi - 1} + \frac{M^2}{(1-\phi)m}(T^{-\phi} - T^{-1})$$
	for some $M_f > 0$, where $T$ denotes the iteration number. Moreover, for a given $\epsilon \in (0,1)$, to guarantee that  $\frac{1}{T} \sum_{k=1}^T \mathbb{E}[\|\nabla f(x_k)\|^2] < \epsilon$, the number of required iterations $T$ is at most $O\left( \epsilon^{-\frac{1}{1-\phi}}\right)$.
\end{theorem}

\section{Numerical Experiments.} \label{sec:exp}

In this section, we compare SMB and SMBi against Adam \citep{Kingma:2015}, and SLS (SGD+Armijo) \citep{Vas:2019}. We have chosen SLS, since it is a recent method that uses stochastic line search with backtracking. We have conducted experiments on multi-class classification problems using neural network models\footnote{The implementations of the models are taken from \url{https://github.com/IssamLaradji/sls}}. Our Python package \texttt{SMB} along with the scripts to conduct our experiments are available online: \url{https://github.com/sibirbil/SMB}

We start our experiments with constant stepsizes for all methods. We should point out that SLS method adjusts the stepsize after each backtracking process and also uses a stepsize reset algorithm between epochs. We refer to this routine as stepsize auto-scheduling. Our numerical experiments show that even without such an auto-scheduling the performances of our methods are on par with SLS. Following the experimental setup in [9], the default setting for hyperparameters of Adam and SLS is used and \(\alpha_0\) has been set to 1 for SLS and 0.001 for Adam. As regards SMB and SMBi, the constant learning rates have been fixed to 0.5, and the constant \(c = 0.1\) as in SLS. Due to the high computational costs of training the neural networks, we report the results of a single run of each method.

\paragraph{MNIST dataset.}  On the MNIST dataset, we have used the one hidden-layer multi-layer perceptron (MLP) of width 1,000.

\begin{figure}[H]
	\begin{subfigure}{\textwidth}
		\includegraphics[width=\textwidth]{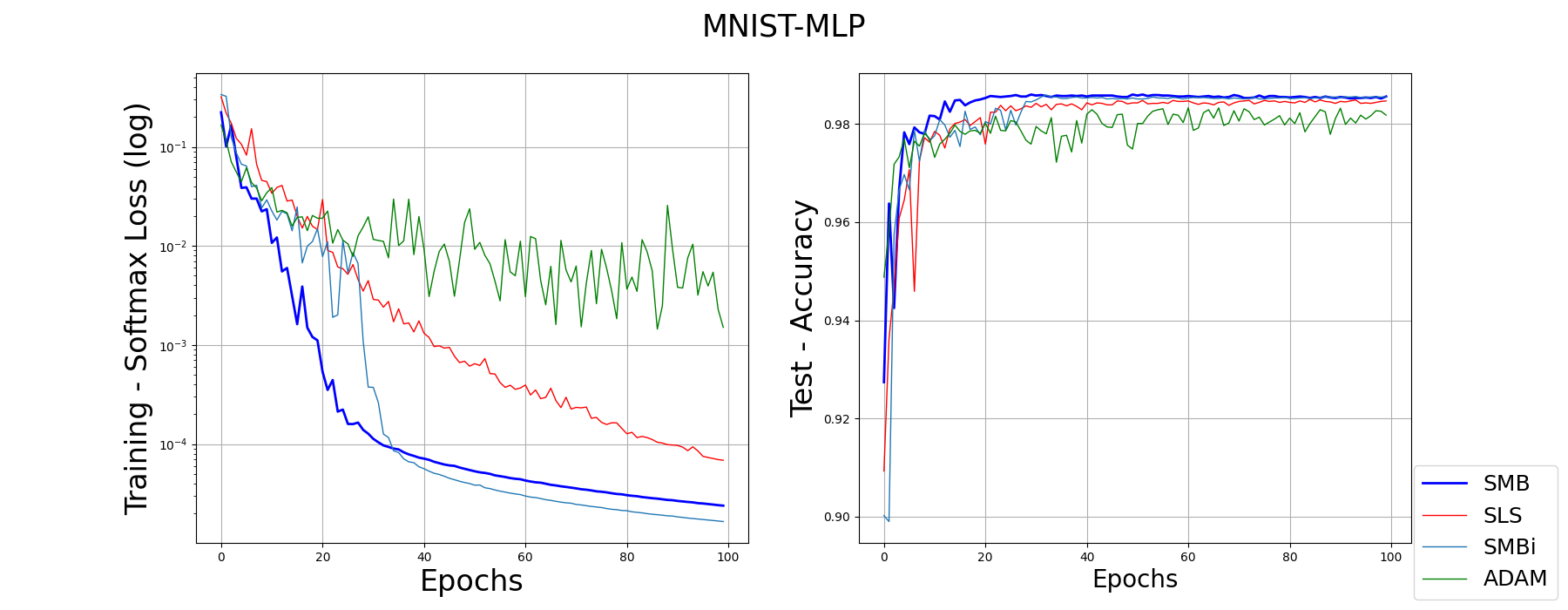}
		\caption*{Training and Test Losses}
	\end{subfigure}%
	\hfill
	\begin{subfigure}{\textwidth}
		\includegraphics[width=1\textwidth]{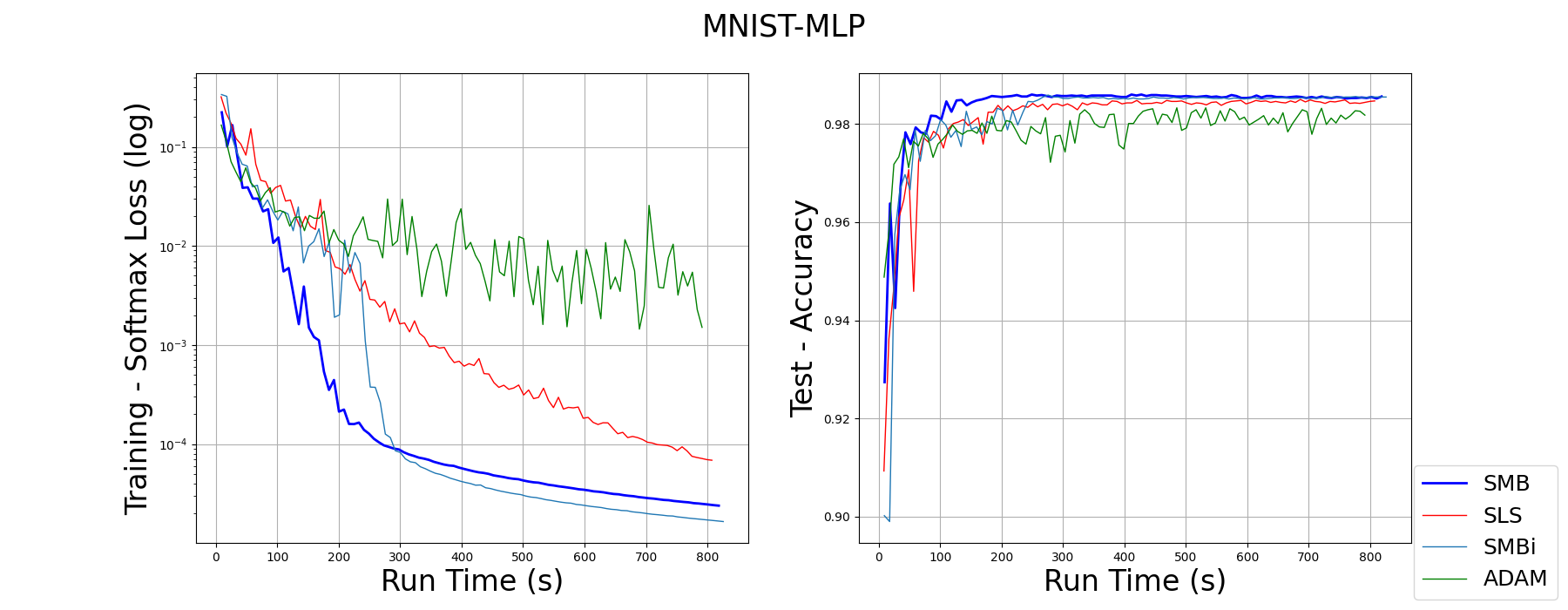}
		\caption*{Training and Test Run Times w.r.t 100 epochs}
	\end{subfigure}
	\caption{Classification on MNIST with an MLP model.}
	\label{fig:mnist}
\end{figure}

In Figure \ref{fig:mnist}, we see the best performances of all four methods on the MNIST dataset with respect to epochs and run time. The run time represents the total time cost of 100 epochs. Even though SMB and SMBi may calculate an extra function value (forward pass) and a gradient (backward pass), we see in this problem that SMB and SMBi achieve the best performance with respect to the run time as well as the number of epochs. More importantly, the generalization performances of SMB and SMBi are also better than the remaining three methods.

It should be pointed out that, in practice, choosing a new independent batch means the SMBi method can construct a model step in two iteration using two batches. This way the computation cost for each iteration is reduced on average with respect to SMB but the model steps can only be taken in half of the iterations in the epoch. As seen in Figure \ref{fig:mnist}, this does not seem to effect the performance in this problem significantly.

\paragraph{CIFAR10 and CIFAR100 datasets.} For the CIFAR10 and CIFAR100 datasets, we have used the standard image-classification architectures ResNet-34 \citep{He:2016} and DenseNet-121 \citep{Huang:2017}. As before, we provide performances of all four methods with respect to epochs and run time. The run times represent the total time cost of 200 epochs.

\begin{figure}[H]
	\centering
	\begin{subfigure}{.5\textwidth}
		\centering
		\includegraphics[width=1\linewidth]{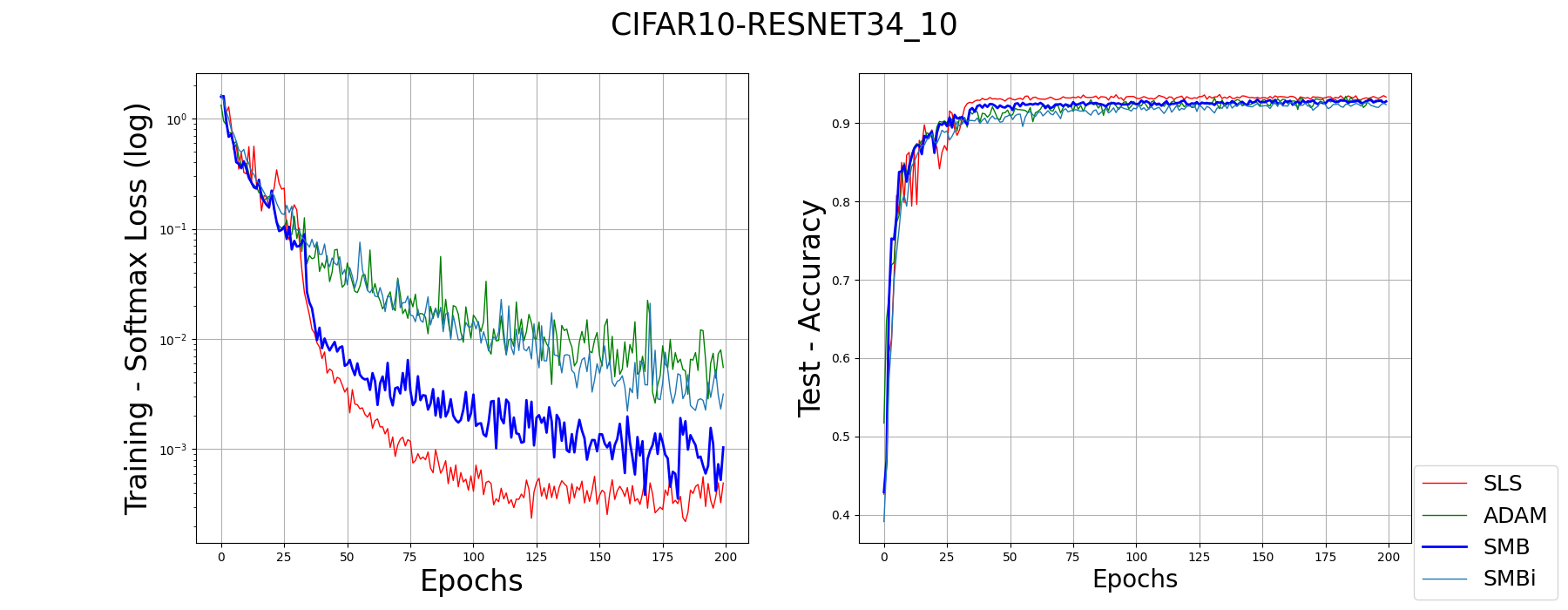}
		\caption*{Training \& Test Losses}
	\end{subfigure}%
	\begin{subfigure}{.5\textwidth}
		\centering
		\includegraphics[width=1\linewidth]{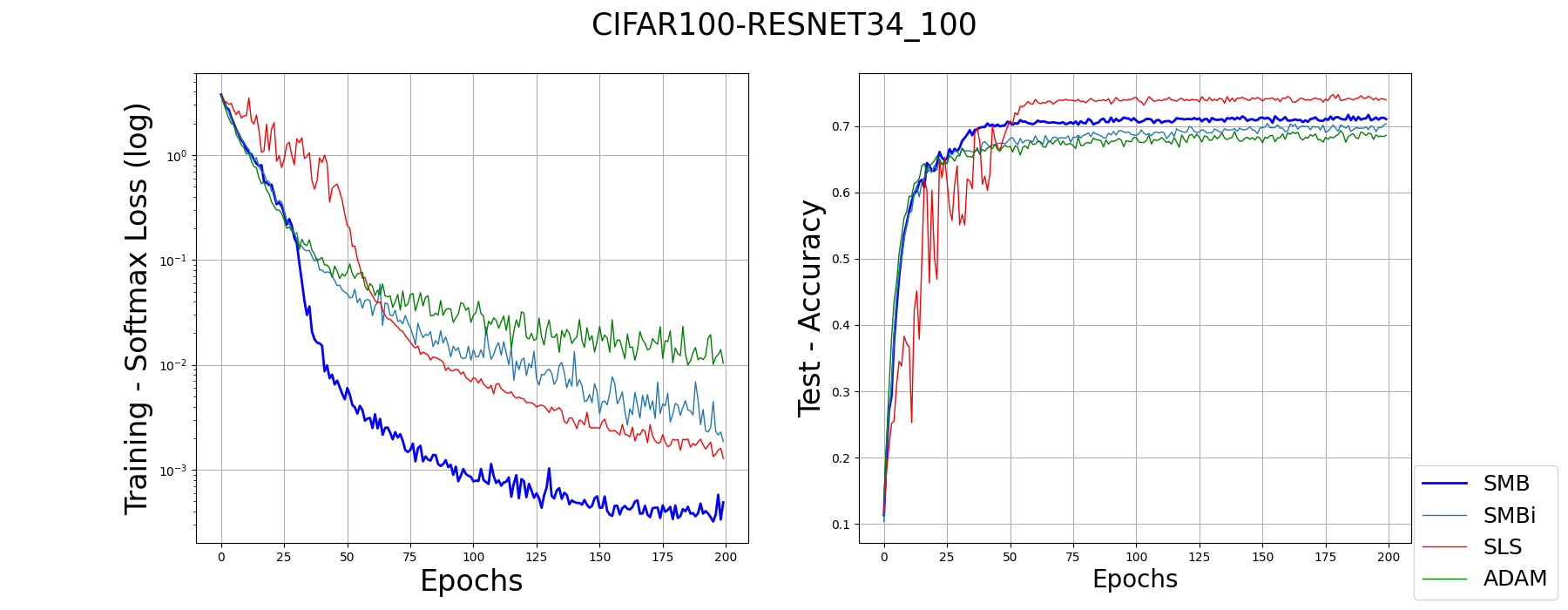}
		\caption*{Training \& Test Losses}
	\end{subfigure}
	\begin{subfigure}{.5\textwidth}
		\centering
		\includegraphics[width=1\linewidth]{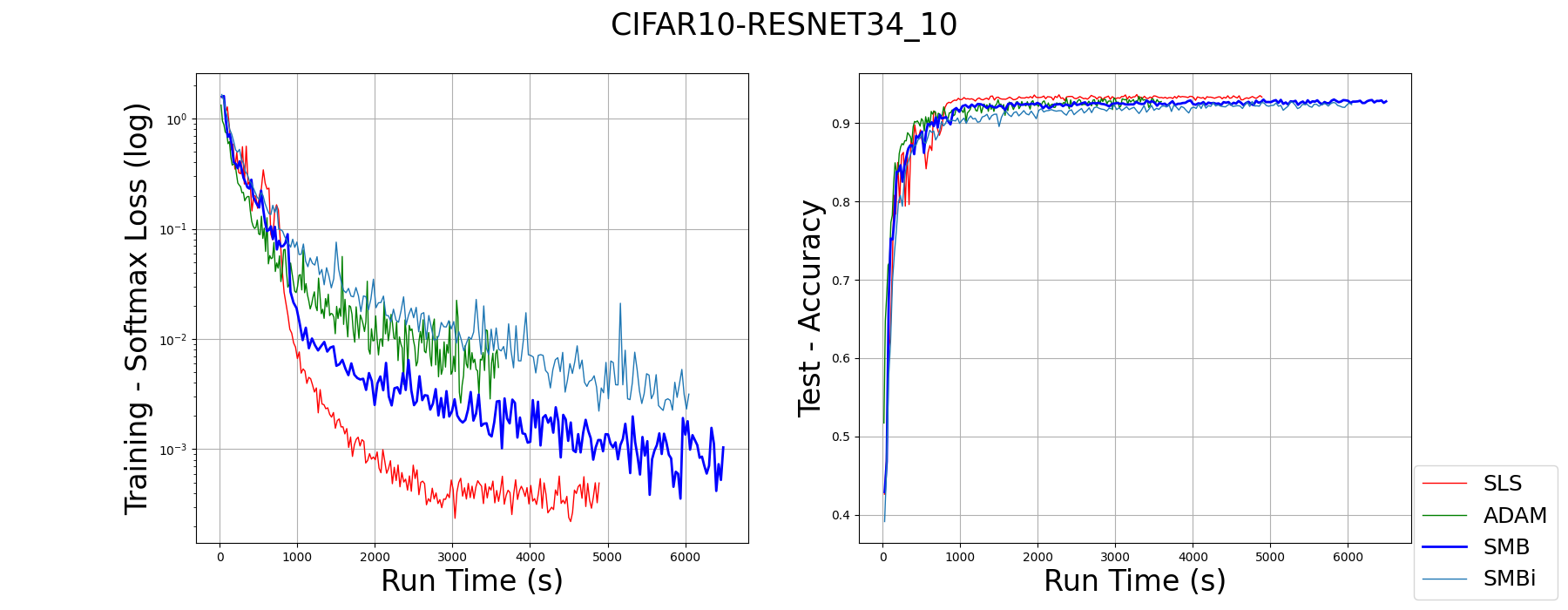}
		\caption*{Training \& Test Run Times w.r.t. 200 epochs}
	\end{subfigure}%
	\begin{subfigure}{.5\textwidth}
		\centering
		\includegraphics[width=1\linewidth]{cifar100-resnet34_100_accuracy.png}
		\caption*{Training \& Test Run Times w.r.t. 200 epochs}
	\end{subfigure}

	\caption{Classification on CIFAR10 (left column) and CIFAR100 (right column) with ResNet-34 model.}
	\label{fig:resnet}
\end{figure}

In Figure \ref{fig:resnet}, we see that on CIFAR10-Resnet34, SMB performs better than Adam algorithm. However, its performance is only comparable to SLS. Even though SMB reaches a lower training loss value in CIFAR100-Resnet34, this advantage does not show in test accuracy.

\begin{figure}[H]
	\centering
	\begin{subfigure}{.5\textwidth}
		\centering
		\includegraphics[width=1\linewidth]{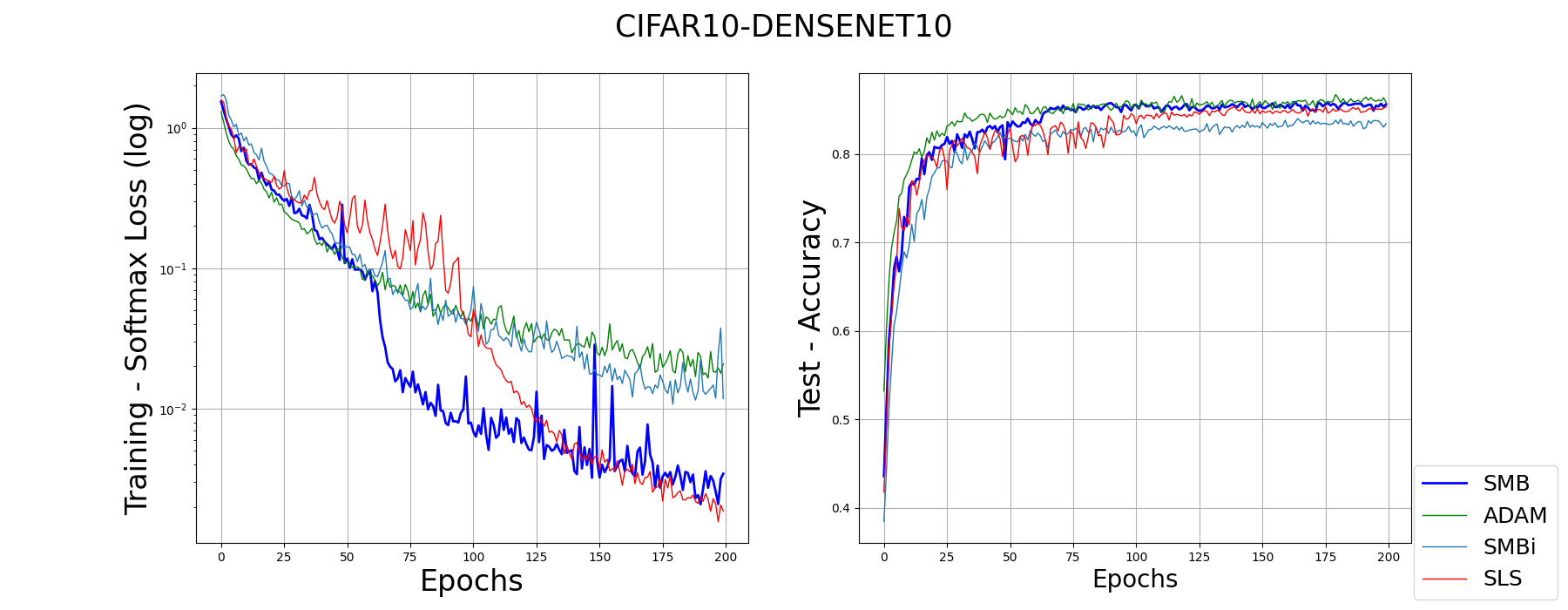}
		\caption*{Training \& Test Losses}
	\end{subfigure}%
	\begin{subfigure}{.5\textwidth}
		\centering
		\includegraphics[width=1\linewidth]{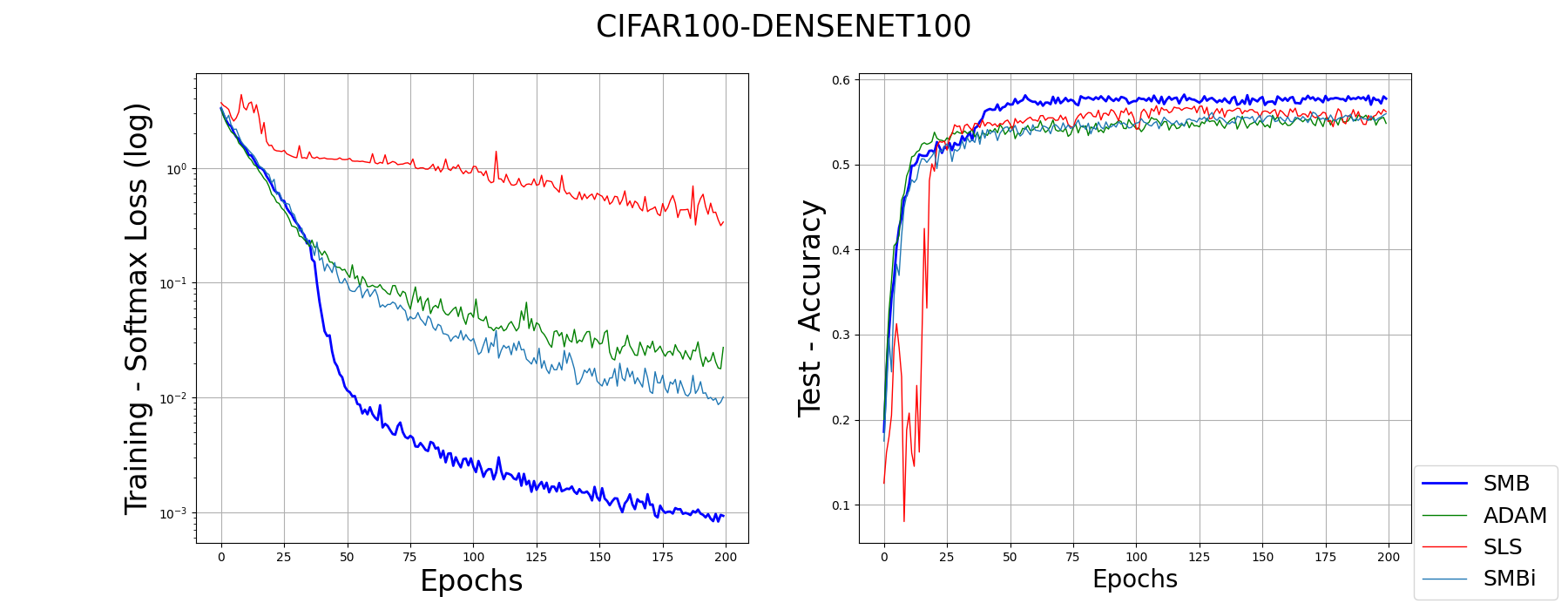}
		\caption*{Training \& Test Losses}
	\end{subfigure}
	\begin{subfigure}{.5\textwidth}
		\centering
		\includegraphics[width=1\linewidth]{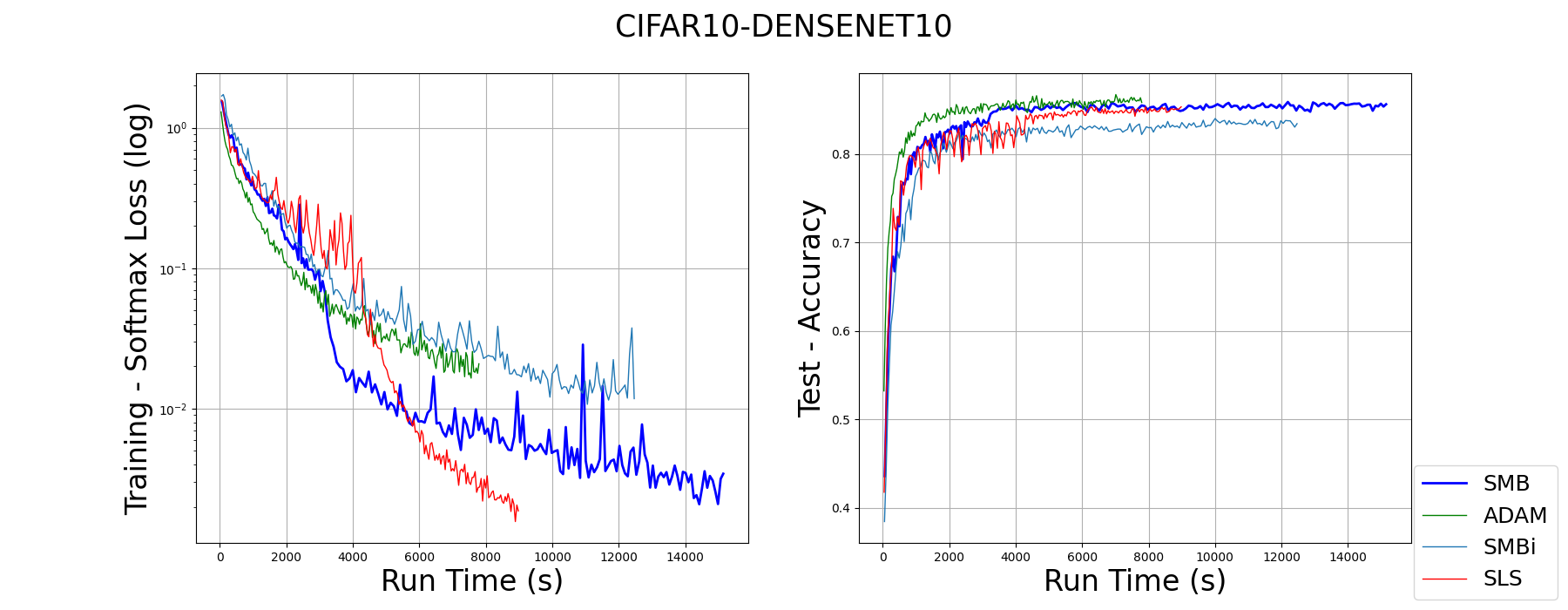}
		\caption*{Training \& Test Run Times w.r.t. 200 epochs}
	\end{subfigure}%
	\begin{subfigure}{.5\textwidth}
		\centering
		\includegraphics[width=1\linewidth]{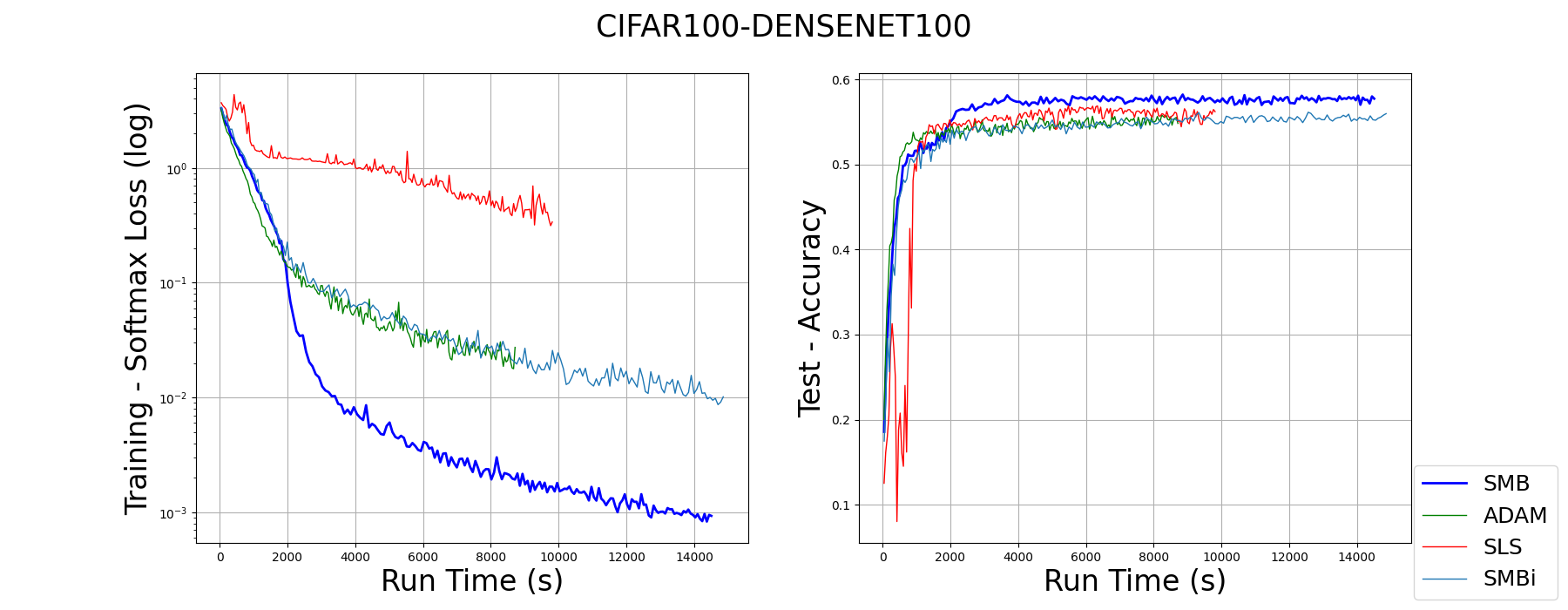}
		\caption*{Training \& Test Run Times w.r.t. 200 epochs}
	\end{subfigure}
	\caption{Classification on CIFAR10 (left column) and CIFAR100 (right column) with Densenet121 model.}
	\label{fig:densenet}
\end{figure}

In Figure \ref{fig:densenet}, we see a comparison of performances of on CIFAR10 and CIFAR100 with DenseNet121. SMB with a constant stepsize outperforms all other optimizers in terms of training error and reaches the best test accuracy on CIFAR100, while showing similar accuracy with ADAM on CIFAR10.

\begin{figure}[H]
	\centering
	\includegraphics[width=1.0\textwidth]{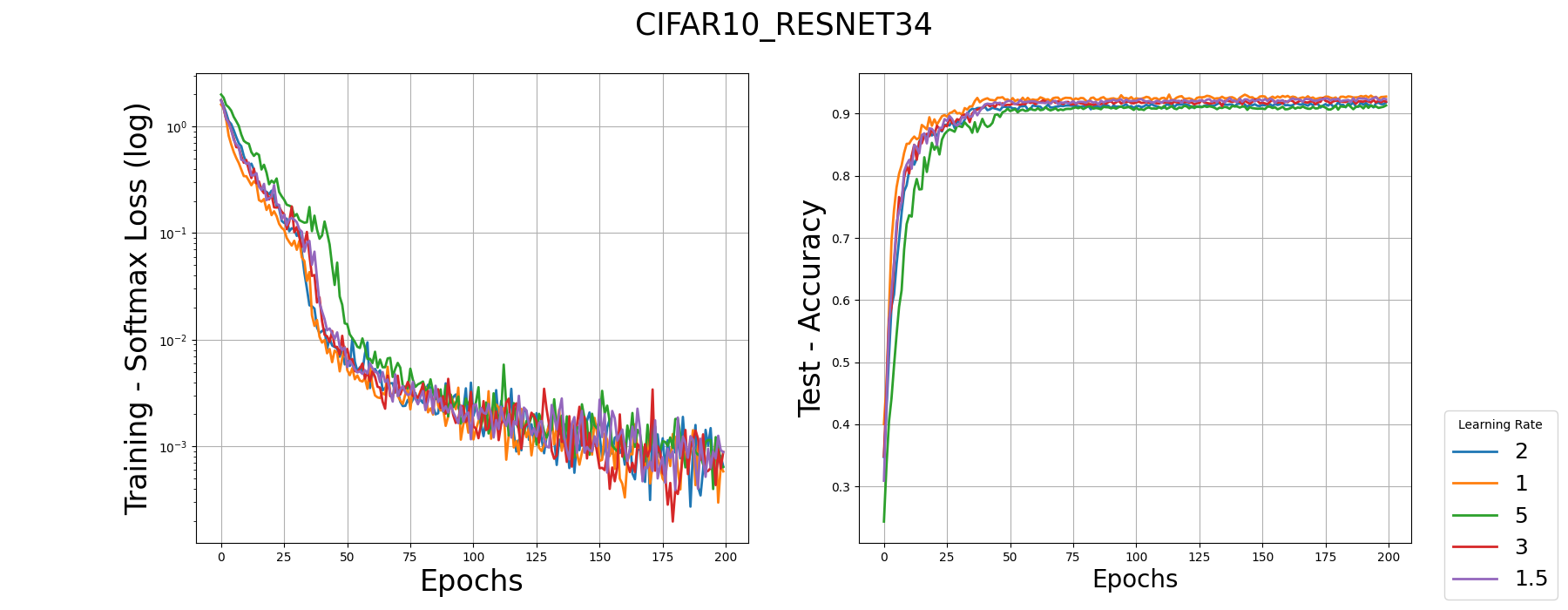}
        \includegraphics[width=1.0\textwidth]{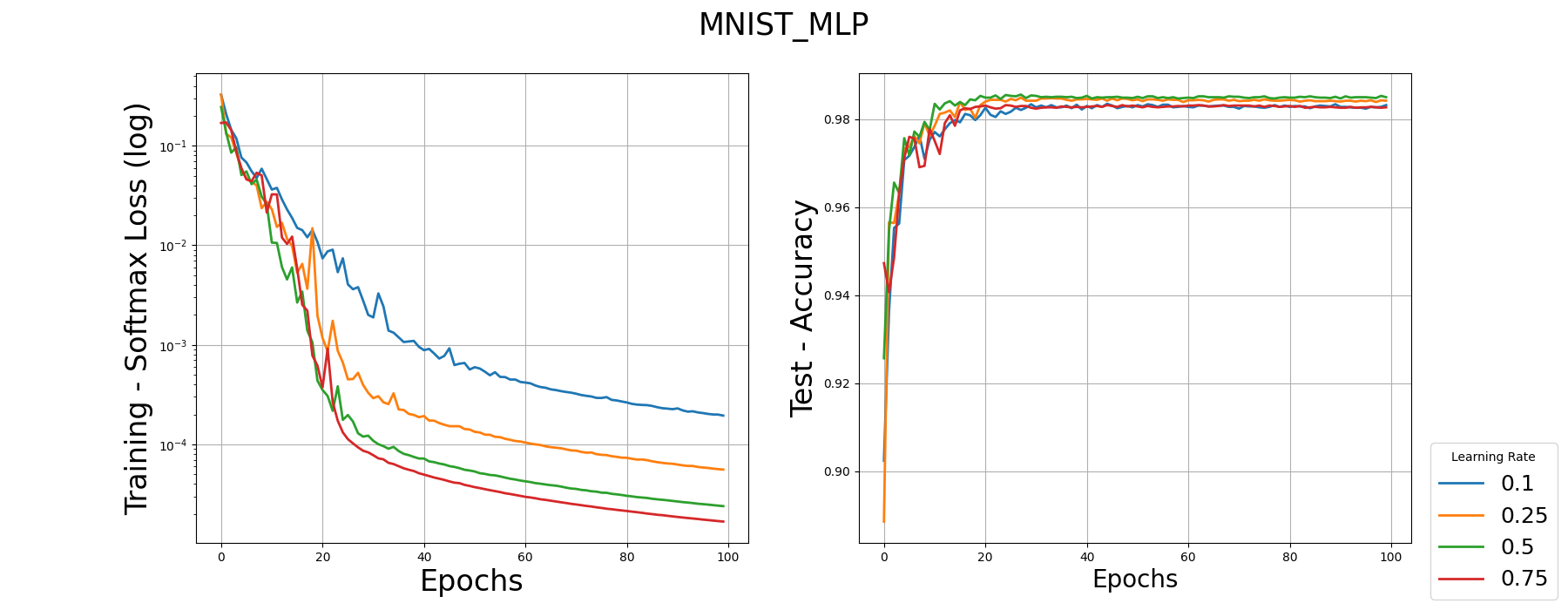}
	\caption{Robustness of SMB under different choices of the learning rate.}
	\label{fig:robust}
\end{figure}

Our last set of experiments are devoted to demonstrating the robustness of SMB. The preliminary results in Figure \ref{fig:robust} show that SMB is robust to the choice of the learning rate, especially in deep neural networks.  This aspect of SMB needs more attention theoretically and experimentally.

\section{Conclusion.} \label{sec:conc}

Stochastic model building (SMB) is a fast alternative to stochastic gradient descent method. The algorithm provides a model building approach that replaces the one-step backtracking in stochastic line search methods. We have analyzed the convergence properties of a modification of SMB by rewriting its model building step as a quasi-Newton update and constructing the scaling matrix with a new independent batch. Our numerical results have shown that SMB converges fast and its performance is insensitive to the selected step length.

In its current state, SMB lacks any internal learning rate adjusting mechanism that could reset the learning rate depending on the progression of the iterations. Our initial experiments show that SMB can greatly benefit from a step length auto-scheduling routine. This is a future work that we will consider. Our convergence rate analysis is given for the alternative algorithm SMBi which can perform competitive against other methods, but consistently underperforms the original SMB method. This begs for a convergence analysis for the SMB method.


\bibliographystyle{apalike}
\bibliography{smb}

\begin{thebibliography}{}

\bibitem[Asi and Duchi, 2019]{Asi2019}
Asi, H. and Duchi, J.~C. (2019).
\newblock The importance of better models in stochastic optimization.
\newblock {\em Proceedings of the National Academy of Sciences},
  116(46):22924--22930.

\bibitem[Balles et~al., 2017]{Balles:2016}
Balles, L., Romero, J., and Hennig, P. (2017).
\newblock Coupling adaptive batch sizes with learning rates.
\newblock In Elidan, G., Kersting, K., and Ihler, A., editors, {\em Proceedings
  of the Thirty-Third Conference on Uncertainty in Artificial Intelligence,
  {UAI} 2017, Sydney, Australia, August 11-15, 2017}. {AUAI} Press.

\bibitem[Bollapragada et~al., 2018]{Bolla:2018}
Bollapragada, R., Byrd, R., and Nocedal, J. (2018).
\newblock Adaptive sampling strategies for stochastic optimization.
\newblock {\em SIAM Journal on Optimization}, 28(4):3312--3343.

\bibitem[Bottou et~al., 2018]{Bottou:2018}
Bottou, L., Curtis, F.~E., and Nocedal, J. (2018).
\newblock Optimization methods for large-scale machine learning.
\newblock {\em SIAM Review}, 60(2):223--311.

\bibitem[Byrd et~al., 2012]{Byrd:2012}
Byrd, R.~H., Chin, G.~M., Nocedal, J., and Wu, Y. (2012).
\newblock Sample size selection in optimization methods for machine learning.
\newblock {\em Mathematical Programming}, 134(1):127--155.

\bibitem[Byrd et~al., 2016]{Byrd:2016}
Byrd, R.~H., Hansen, S.~L., Nocedal, J., and Singer, Y. (2016).
\newblock A stochastic quasi-newton method for large-scale optimization.
\newblock {\em SIAM Journal on Optimization}, 26(2):1008--1031.

\bibitem[Chen et~al., 2023]{chen:2023}
Chen, Y.-L., Na, S., and Kolar, M. (2023).
\newblock Convergence analysis of accelerated stochastic gradient descent under
  the growth condition.
\newblock {\em Mathematics of Operations Research}.
\newblock Available online: https://doi.org/10.1287/moor.2021.0293.

\bibitem[Defazio et~al., 2014]{defazio:2014}
Defazio, A., Bach, F., and Lacoste-Julien, S. (2014).
\newblock {SAGA}: a fast incremental gradient method with support for
  non-strongly convex composite objectives.
\newblock In {\em Proceedings of the 27th International Conference on Neural
  Information Processing Systems - Volume 1}, NIPS'14, page 1646–1654,
  Cambridge, MA, USA. MIT Press.

\bibitem[He et~al., 2016]{He:2016}
He, K., Zhang, X., Ren, S., and Sun, J. (2016).
\newblock Deep residual learning for image recognition.
\newblock In {\em 2016 IEEE Conference on Computer Vision and Pattern
  Recognition (CVPR)}, pages 770--778.

\bibitem[Huang et~al., 2017]{Huang:2017}
Huang, G., Liu, Z., Maaten, L. V.~D., and Weinberger, K.~Q. (2017).
\newblock Densely connected convolutional networks.
\newblock In {\em 2017 IEEE Conference on Computer Vision and Pattern
  Recognition (CVPR)}, pages 2261--2269, Los Alamitos, CA, USA. IEEE Computer
  Society.

\bibitem[Khaled and Richt{\'a}rik, 2020]{khaled:2020}
Khaled, A. and Richt{\'a}rik, P. (2020).
\newblock Better theory for {SGD} in the nonconvex world.
\newblock {\em ArXiv}, abs/2002.03329.

\bibitem[Kingma and Ba, 2015]{Kingma:2015}
Kingma, D.~P. and Ba, J. (2015).
\newblock Adam: {A} method for stochastic optimization.
\newblock In Bengio, Y. and LeCun, Y., editors, {\em 3rd International
  Conference on Learning Representations, {ICLR} 2015, San Diego, CA, USA, May
  7-9, 2015, Conference Track Proceedings}.

\bibitem[Liuzzi et~al., 2022]{liuzzi:2022}
Liuzzi, G., Palagi, L., and Seccia, R. (2022).
\newblock Convergence under {L}ipschitz smoothness of ease-controlled random
  reshuffling gradient algorithms.
\newblock {\em ArXiv}, abs/2212.01848.

\bibitem[Mahsereci and Hennig, 2017]{Hennig:2017}
Mahsereci, M. and Hennig, P. (2017).
\newblock Probabilistic line searches for stochastic optimization.
\newblock {\em The Journal of Machine Learning Research}, 18(1):4262--4320.

\bibitem[Malinovsky et~al., 2022]{malinovsky:2022}
Malinovsky, G., Mishchenko, K., and Richt{\'a}rik, P. (2022).
\newblock Server-side stepsizes and sampling without replacement provably help
  in federated optimization.
\newblock {\em ArXiv}, abs/2201.11066.

\bibitem[Mokhtari and Ribeiro, 2014]{Mokhtari:2014}
Mokhtari, A. and Ribeiro, A. (2014).
\newblock {RES}: Regularized stochastic {BFGS} algorithm.
\newblock {\em IEEE Transactions on Signal Processing}, 62(23):6089--6104.

\bibitem[Mutschler and Zell, 2020]{Muts:2020}
Mutschler, M. and Zell, A. (2020).
\newblock Parabolic approximation line search for {DNNs}.
\newblock In Larochelle, H., Ranzato, M., Hadsell, R., Balcan, M., and Lin, H.,
  editors, {\em Advances in Neural Information Processing Systems}, volume~33,
  pages 5405--5416. Curran Associates, Inc.

\bibitem[{\"O}ztoprak and Birbil, 2018]{Oztoprak:2017}
{\"O}ztoprak, F. and Birbil, {\c{S}}.~{\.I}. (2018).
\newblock An alternative globalization strategy for unconstrained optimization.
\newblock {\em Optimization}, 67(3):377--392.

\bibitem[Paquette and Scheinberg, 2020]{Paq:2020}
Paquette, C. and Scheinberg, K. (2020).
\newblock A stochastic line search method with expected complexity analysis.
\newblock {\em SIAM Journal on Optimization}, 30(1):349--376.

\bibitem[Roux et~al., 2012]{roux:2012}
Roux, N.~L., Schmidt, M., and Bach, F. (2012).
\newblock A stochastic gradient method with an exponential convergence rate for
  finite training sets.
\newblock In {\em Proceedings of the 25th International Conference on Neural
  Information Processing Systems - Volume 2}, NIPS'12, page 2663–2671, Red
  Hook, NY, USA. Curran Associates Inc.

\bibitem[Schraudolph et~al., 2007]{Schrau:2007}
Schraudolph, N.~N., Yu, J., and Günter, S. (2007).
\newblock A stochastic quasi-newton method for online convex optimization.
\newblock In Meila, M. and Shen, X., editors, {\em Proceedings of the Eleventh
  International Conference on Artificial Intelligence and Statistics}, volume~2
  of {\em Proceedings of Machine Learning Research}, pages 436--443, San Juan,
  Puerto Rico. PMLR.

\bibitem[Tadi{\'c}, 1997]{tadic:1997}
Tadi{\'c}, V. (1997).
\newblock Stochastic gradient algorithm with random truncations.
\newblock {\em European Journal of Operational Research}, 101(2):261--284.

\bibitem[Vaswani et~al., 2019]{Vas:2019}
Vaswani, S., Mishkin, A., Laradji, I., Schmidt, M., Gidel, G., and
  Lacoste-Julien, S. (2019).
\newblock {\em Painless stochastic gradient: interpolation, line-search, and
  convergence rates}.
\newblock Curran Associates Inc., Red Hook, NY, USA.

\bibitem[Wang et~al., 2017]{Wang:2017}
Wang, X., Ma, S., Goldfarb, D., and Liu, W. (2017).
\newblock Stochastic quasi-newton methods for nonconvex stochastic
  optimization.
\newblock {\em SIAM Journal on Optimization}, 27(2):927--956.

\end{thebibliography}

\clearpage

\appendix

\noindent{\large \bf APPENDIX}

\section*{Proof of Theorem \ref{thm:wang}} \label{app:proof}

First we show that the SMBi step for each parameter group $p$ can be expressed as a special quasi-Newton update. For brevity, let us use $s_k$, $s_k^t$, $g_k$, $g_k^t$, and $y_k$ instead of $s_{k,p}$, $s_{k,p}^t$, $g_{k,p}$, $g_{k,p}^t$, and $y_{k,p}$, respectively. Recalling the definitions of $\theta$ and $\delta$ given in (\ref{theta_delta}), observe that
\[
	2\delta = \|s_k^t\|\|y_k\|+\frac{1}{\eta} \|s_k^t\|\|g_k\| - y_k^\top s_k^t
	= \alpha_k \left(  \|g_k\|\|y_k\|+\frac{1}{\eta} \|g_k\|^2 + y_k^\top g_k\right) = \alpha_k \sigma,
\]
and
\[
	\theta = \left(y_k^\top s_k^t + 2\delta\right)^2-\|s_k^t\|^2\|y_k\|^2
	= \alpha_k^2 (\sigma - y_k^\top g_k )^2  - \alpha_k^2 \|g_k\|^2\|y_k\|^2
	= \alpha_k^2  (\beta^2 - \|g_k\|^2\|y_k\|^2) = \alpha_k^2 \gamma,
\]
where
$$\sigma = \|g_k\|\|y_k\|+\frac{1}{\eta} \|g_k\|^2 + y_k^\top g_k, \
	\beta = \sigma - y_k^\top g_k, \mbox{ and }
	\gamma = (\beta^2 - \|g_k\|^2\|y_k\|^2).$$

Therefore, we have
\[
	c_g(\delta) g_k = -\frac{\|s_k^t\|^2}{2\delta} g_k = -\frac{\alpha_k^2 \|g_k\|^2}{\alpha_k \sigma \gamma}  \gamma g_k = -\alpha_k \frac{\|g_k\|^2}{\sigma \gamma}  \gamma g_k,
\]
\begin{align*}
	c_y(\delta) y_k & = -\frac{\|s_k^t\|^2}{2\delta\theta}[-(y_k^\top s_k^t +
	2\delta)(s_k^t)^\top g_k + \|s_k^t\|^2 y_k^\top g_k] y_k                                                          \\
	                & =  -\frac{\|g_k\|^2}{\alpha_k\sigma\gamma} y_k [\alpha_k^2(\sigma - y_k^\top g_k) g_k^\top g_k+
	\alpha_k^2 \|g_k\|^2 y_k^\top g_k]                                                                                \\
	                & = -\alpha_k \frac{\|g_k\|^2}{\sigma\gamma} [\beta y_k g_k^\top + \|g_k\|^2  y_k y_k^\top] g_k,
\end{align*}
and
\begin{align*}
	c_s(\delta) s^t_k & = -\frac{\|s_k^t\|^2}{2\delta\theta}[-(y_k^\top s_k^t + 2\delta)y_k^\top g_k
	+ \|y_k\|^2(s_k^t)^\top g_k]  s^t_k                                                                               \\
	                  & = -\frac{\|g_k\|^2}{\alpha_k\sigma\gamma} (- \alpha_k) g_k [-\alpha_k (\sigma - y_k^\top g_k)
	y_k^\top g_k - \alpha_k \|y_k\|^2 g_k^\top g_k]                                                                   \\
	                  & = -\alpha_k \frac{\|g_k\|^2}{\sigma\gamma}[\beta g_k y_k^\top +  \|y_k\|^2 g_k g_k^\top] g_k.
\end{align*}
Now, it is easy to see that
\begin{align*}
	s_k & = c_g(\delta) g_k + c_y(\delta) y_k + c_s(\delta) s^t_k   \\
	    & = -\alpha_k \frac{\|g_k\|^2}{\sigma\gamma}\left[ \gamma I
		+ \beta y_k g_k^\top + \|g_k\|^2 y_k y_k^\top
		+ \beta g_k y_k^\top + \|y_k\|^2 g_k g_k^\top \right] g_k.
\end{align*}
Thus, for each parameter group $p$, we define
\begin{equation}
	\label{eq:QN}
	H_{k,p} =    \frac{\|g_{k,p}\|^2}{\sigma_p\gamma_p}\left[ \gamma_p I
	+ \beta_p y_{k,p} g_{k,p}^\top + \|g_{k,p}\|^2 y_{k,p} y_{k,p}^\top
	+ \beta_p g_{k,p} y_{k,p}^\top + \|y_{k,p}\|^2 g_{k,p} g_{k,p}^\top \right],
\end{equation}\
where
$$\sigma_p = \|g_{k,p}\|\|y_{k,p}\|+\frac{1}{\eta} \|g_{k,p}\|^2 + y_{k,p}^\top g_{k,p}, \
	\beta_p = \sigma_p - y_{k,p}^\top g_{k,p}, \mbox{ and }
	\gamma_p = (\beta_p^2 - \|g_{k,p}\|^2\|y_{k,p}\|^2).$$

Now, assuming that we have the parameter groups $\{p_1, \dots, p_n\}$, the SMB steps can be expressed as a quasi-Newton update given by
\[
	x_{k+1} = x_k -\alpha_k H_k g_k,
\]
where
\[
	H_k = \begin{cases}
		I,                                         & \mbox{if the Armijo condition is satisfied;} \\
		\mbox{diag}(H_{k,p_1}, \ldots, H_{k,p_n}), & \mbox{otherwise.}
	\end{cases}
\]
Here, $I$ denotes the identity matrix, and $\mbox{diag}(H_{k,p_1}, \ldots, H_{k,p_n})$ denotes the block diagonal matrix with the blocks $H_{k,p_1}, \ldots, H_{k,p_n}$.

We next show that the eigenvalues of the matrices $H_k$, $k \geq 1$, are bounded from above and below uniformly which is, of course, obvious when $H_k = I$. Using the Sherman-Morrison formula twice, one can see that for each parameter group $p$, the matrix $H_{k,p}$ is indeed the inverse of the positive semidefinite matrix
$$B_{k,p} = \frac{1}{\|g_{k,p}\|^2}(\sigma_p I - g_{k,p} y_{k,p}^\top - y_{k,p} g_{k,p}^\top),$$
and hence, it is also positive semidefinite. Therefore, it is enough to show the boundedness of the eigenvalues of $B_{k,p}$ uniformly on $k$ and $p$.

Since $g_{k,p} y_{k,p}^\top + y_{k,p} g_{k,p}^\top$ is a rank two matrix, $\sigma_p / \|g_{k,p}\|^2$ is an eigenvalue of $B_{k,p}$ with multiplicity $n-2$. The remaining extreme eigenvalues are
\[
	\lambda_{max}(B_{k,p}) = \frac{1}{\|g_{k,p}\|^2}(\sigma_p + \|g_{k,p}\| \|y_{k,p}\| - y_{k,p}^\top g_{k,p}) \ \ \mbox{  and  } \ \
	\lambda_{min}(B_{k,p}) = \frac{1}{\|g_{k,p}\|^2}(\sigma_p - \|g_{k,p}\| \|y_{k,p}\| - y_{k,p}^\top g_{k,p})
\]
with the corresponding eigenvectors $\|y_{k,p}\| g_{k,p} + \|g_{k,p}\| y_{k,p}$ and $\|y_{k,p}\| g_{k,p} - \|g_{k,p}\| y_{k,p}$, respectively.

Observe that,
\begin{align*}
	\lambda_{min}(B_{k,p}) & = \frac{\sigma_p - \|g_{k,p}\| \|y_{k,p}\| - y_{k,p}^\top g_{k,p}}{\|g_{k,p}\|^2}                                                                 \\
	                       & = \frac{\|g_{k,p}\| \|y_{k,p}\| + \eta^{-1} \|g_{k,p}\|^2 + y_{k,p}^\top g_{k,p} - \|g_{k,p}\| \|y_{k,p}\| - y_{k,p}^\top g_{k,p}}{\|g_{k,p}\|^2} \\
	                       & = \frac{\eta^{-1} \|g_{k,p}\|^2}{\|g_{k,p}\|^2} = \frac{1}{\eta} > 1.
\end{align*}
Thus, the smallest eigenvalue $B_{k,p}$ is bounded away from zero uniformly on $k$ and $p$.

Now, by our assumption of Lipschitz continuity of the gradients, for any $x,y \in \mathbb{R}^n$ and $\xi_k$, we have
\[
	\|g(x, \xi_{k}) - g(y, \xi_{k})\| \leq L \|x - y\|.
\]
Thus, observing that $\|y_{k,p}\| = \|g_{k,p}^t - g_{k,p}\| \leq L \|x_{k,p}^t - x_{k,p}\| \leq \alpha_k L \|g_{k,p}\|$, we have
\begin{align*}
	\lambda_{max}(B_{k,p}) & = \frac{\sigma_p + \|g_{k,p}\| \|y_{k,p}\| - y_{k,p}^\top g_{k,p}}{\|g_{k,p}\|^2}                                                                  \\
	                       & = \frac{\|g_{k,p}\| \|y_{k,p}\| + \eta^{-1} \|g_{k,p}\|^2 + y_{k,p}^\top g_{k,p} + \|g_{k,p}\| \|y_{k,p}\| - y_{k,p}^\top g_{k,p}}{\|g_{k,p}\|^2}  \\
	                       & = \frac{2 \|g_{k,p}\| \|y_{k,p}\| + \eta^{-1} \|g_{k,p}\|^2}{\|g_{k,p}\|^2} \leq 2 L \alpha_k +  \frac{1}{\eta} \leq 2 L \alpha_{max} + \eta^{-1}.
\end{align*}
This implies that the eigenvalues of $H_{k,p} = B_{k,p}^{-1}$ are bounded below by $1/(\eta^{-1} + 2 L \alpha_{max})$ and bounded above by 1 uniformly on $k$ and $p$. This result, together with our assumptions, shows that steps of the SMBi algorithm satisfy the conditions of Theorem 2.10 in \citep{Wang:2017} with $\underline{\kappa} = 1/(\eta^{-1} + 2 L \alpha_{max})$ and $\overline{\kappa} = 1$ and Theorem \ref{thm:wang} follows as a corollary.


\end{document}